\documentclass{article}
\usepackage{amssymb}
\usepackage{xcolor}
\usepackage{xspace}
\usepackage{amsmath}
\usepackage{enumitem}
\usepackage{graphicx} 
\usepackage{booktabs}
\usepackage{multirow}
\usepackage{multicol}
\usepackage{makecell}
\usepackage[flushleft]{threeparttable}
\newcommand{\ours}{DexCtrl\xspace}

\usepackage[preprint]{corl_2025} % Uncomment for pre-prints (e.g., arxiv); This is like ``final'', but will remove the CORL footnote.

\title{DexCtrl: Towards Sim-to-Real Dexterity with Adaptive Controller Learning}

\author{
  Shuqi Zhao, Ke Yang, Yuxin Chen, Chenran Li, Yichen Xie, \\
  \textbf{$^{\dagger}$Xiang Zhang, $^{\dagger}$Changhao Wang, $^{\dagger}$Masayoshi Tomizuka}\\
  Department of Mechanical Engineering\\
  University of California Berkeley, 
  United States\\
}

\begin{document}
\maketitle

%===============================================================================

\begin{figure}[h!]
  \centering
  \vspace{-20pt}
   \includegraphics[width=\textwidth]{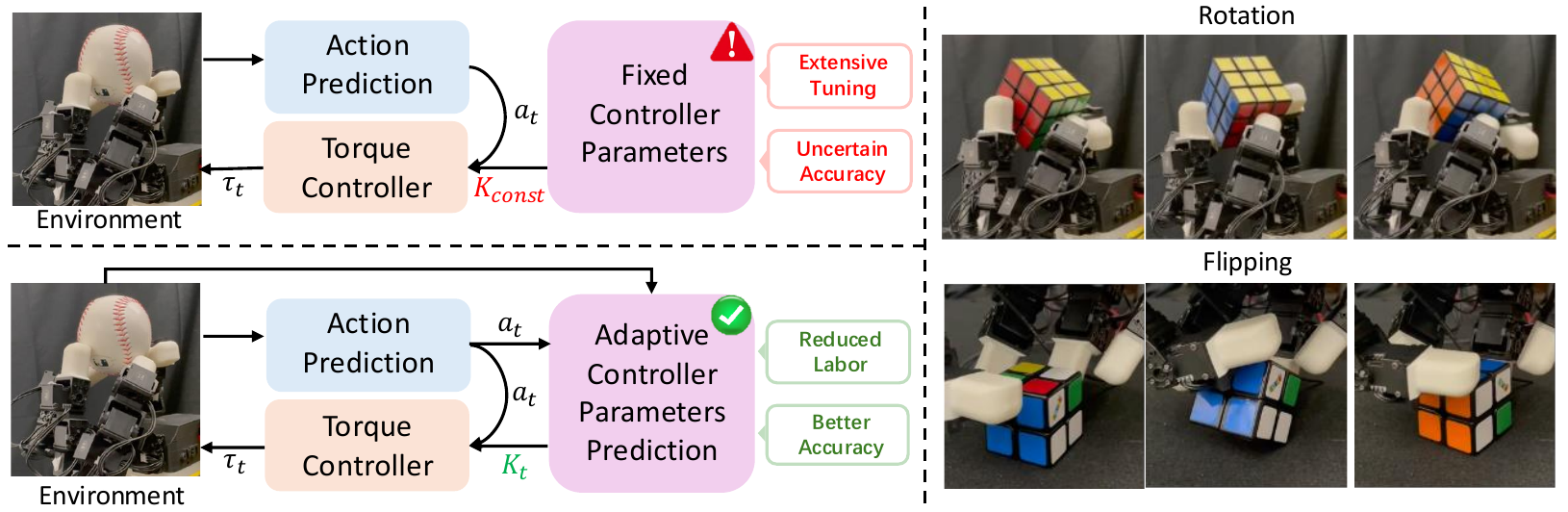}
   \vspace{-18pt}
   \caption{Compared to previous work with only action prediction (upper left), \ours (lower left) jointly predicts both action and control parameters, significantly reducing human labor for tuning and achieving better performance on two contact-rich manipulation tasks: rotation and flipping.}
   \label{fig:teaser}
   \vspace{-10pt}
\end{figure}

\begin{abstract}
    Dexterous manipulation has seen remarkable progress in recent years, with policies capable of executing many complex and contact-rich tasks in simulation. However, transferring these policies from simulation to real world remains a significant challenge. One important issue is the mismatch in low-level controller dynamics, where identical trajectories can lead to vastly different contact forces and behaviors when control parameters vary. Existing approaches often rely on manual tuning or controller randomization, which can be labor-intensive, task-specific, and introduce significant training difficulty. In this work, we propose a framework that jointly learns actions and controller parameters based on the historical information of both trajectory and controller. This adaptive controller adjustment mechanism allows the policy to automatically tune control parameters during execution, thereby mitigating the sim-to-real gap without extensive manual tuning or excessive randomization. Moreover, by explicitly providing controller parameters as part of the observation, our approach facilitates better reasoning over force interactions and improves robustness in real-world scenarios. Experimental results demonstrate that our method achieves improved transfer performance across a variety of dexterous tasks involving variable force conditions.
\end{abstract}


%===============================================================================

\section{Introduction}
Dexterity is a core component of human manipulation and has long posed a significant challenge in robotics research. 
Beyond their strong performance in grasping~\citep{zhang2024dexgraspnet, yin2025geometric, ye2023learning,yang2024anyrotate}, dexterous manipulation policies has shown capabilities in handling various contact-rich manipulation tasks such as rotating objects~\citep{chen2023visual, qi2023hand,qi2023general,wang2024lessons,qi2025simple}, playing the piano~\citep{zakka2023robopianist,qian2024pianomime}, and using various tools~\citep{yin2025dexteritygen,liu2025dextrack}.
%
% These accomplishments highlight the significant potential of dexterous hands for complex manipulation tasks. 
Despite these progress, transferring dexterous policies from simulation to the real world still remains a critical challenge.
Currently, many manipulation policies have addressed the sim-to-real gap from various aspects, such as introducing random noise to the observed proprioceptive information or applying random forces to objects to enhance output trajectory robustness.

However, we here identify that one important issue has received relatively little attention in previous work: the discrepancies between robot controllers in simulation and the real world.
% (one biggest sim-to-real gap is controller: causes some problems.)
Since the final command sent to the robot is the motor torque computed from both the trajectory and control parameters, failing to explicitly consider robot controller gap still results in a discrepancy between simulated and real-world performance.
Nowadays, people try to bridge this gap by manually tuning controller parameters, which compares the robot’s trajectory outputs between simulation and the real world to match the final performance~\citep{chen2023visual, yu2024hand}.
% (people doesn't focus on controller alignment! don't define this as a method.)
Additionally, mild randomization of controller parameters during training has also been widely used to enhance policy robustness against sim-to-real discrepancies~\citep{qi2023general, yin2025dexteritygen}.
% (specifically targeted on control)
%
However, manual tuning and randomization can both ultimately lead to task failure in real-world deployments: randomization leads to a substantial increase in training difficulty, and manually tuning control parameters often fails to achieve the necessary precision for successful task execution.
Moreover, both of them require extensive effort in adjusting controller parameters and tuning randomization hyperparameters, significantly increasing human labor.
Fundamentally, current solutions to this problem have relied merely on extensive attempts based on prior human experiences, rather than a truly principled and automatic approach.

In our work, as shown in Figure~\ref{fig:teaser}, we propose a novel method, \ours, that adaptively adjusts the controller parameters based on historical information, thus narrowing the sim-to-real gap. 
% (address controller issues)
% (sim real robot controller mapping, human labor, task specifical, others ignore this, simulation can obtain this)
Concretely, \ours model learns to output both actions and controller parameters for each time step based on previous desired and actual joint trajectories as well as the corresponding controller parameters within a time window.
By doing so, \ours can automatically adjust the controller behavior in a close-loop manner, bypassing the complicated procedure of manually tuning parameters, with the assurance of strong adaptability to real-world scenarios.
Besides, it alleviates the policy exploration difficulty introduced by randomization because \ours directly obtains controller information in observation, leading to better capture of force information.
% (Maybe? Is it true? Do we need experiments for the contact and force part?).
% (besides controller benefits, sim-to-real factors can be reduced)
% Additionally, our method can deal with force changes more quickly and smoothly because adaptive controller enables better and smoother converge of current trajectories.(If I can make some demos?)
%
Overall, the contributions of our work are as follows:
\begin{itemize}
    \item We identify the mismatch of robot controllers as a critical factor in the sim-to-real gap and propose a novel method to adjust the control parameters adaptively. 
    % We identify controller gap as a critical factor in sim-to-real problem and propose a novel method that effectively narrows the sim-to-real gap by jointly obtaining actions and controller parameters, mitigating the influence of controller mismatch.
    \item We design a simple and elegant framework to jointly obtain actions and controller parameters based on historical information, which can offer better adaptivity to force variation.
    % \item We propose a novel method to adaptively adjust control parameters based on historical information at each step, which proves to offer better adaptivity to force variation.
    \item Extensive experiments on two different tasks show our method can significantly outperform baselines in both simulation and the real world, along with thorough analysis. 
    % showing the importance of control parameters in force-sensitive tasks.
\end{itemize}

\section{Problem Statements}
\textbf{Manipulation Tasks} Our method primarily focuses on enhancing the sim-to-real performance of dexterous manipulation tasks by bridging the control gap. To validate our approach, we implement two challenging dexterous manipulation tasks that involve contact between objects, hands, and environments: in-hand object rotation and flipping.
The goal of the in-hand rotation task is to rotate an object using the fingertips along a specific axis without dropping it, while the goal of the flipping task is to flip an object on a table along a designated axis. Both tasks exemplify the contact-rich behaviors characteristic of dexterous manipulation, where the sim-to-real gap significantly impacts real-world performance.

\textbf{Robot Controller} Our method is primarily designed for policies operating under joint torque control. We use the LEAP hand as an example, which has 16 degrees of freedom (DOF).
The torque controller takes in the robot hand desired joint trajectories $[q^{d}, \dot{q}^{d}] \in \mathbb{R}^{32}$, and robot current trajectories $[q^{c}, \dot{q}^{c}] \in \mathbb{R}^{32}$, and computes the corresponding joint torque $\tau \in \mathbb{R}^{16}$ as follows:
\begin{equation}
    \tau = K_P(q^{d} - q^{c}) + K_D(\dot{q}^{d} - \dot{q}^{c})
\label{eq:controller}
\end{equation}
We assume $K_P$, $K_D$ are diagonal for simplicity, and define $K = \{K_P, K_D\}\in \mathbb{R}^{32}$ as the collection of controller parameters, representing the robot stiffness and damping matrices, respectively.
As shown in Eq.~\ref{eq:controller}, the torque output is directly modulated by the choice of $K$, which necessitates careful tuning of these parameters.
In particular, besides control parameters $K$ that directly determine the actual torque values, increasing stiffness $K_P$ reduces steady-state error but may induce oscillations, while increasing damping $K_D$ suppresses overshoot but can amplify high-frequency noise from a dynamic perspective.
In policies without adaptive control mechanisms, the controller parameters are fixed, and only the desired trajectory $q^d_t$ is predicted by the policy.
Different from previous work, \ours also predicts controller parameters $K_t$ along with $q^d_t$ at each time step, enabling simultaneous controller adjustment at each time step to meet force requirements and ensure smooth trajectories.
Desired velocities $\dot{q}^d$ are set as zero in both our method and previous work.
% (flatten these goals)
\section{Methods}
\label{sec:method}

\begin{figure}[t]
  \centering
   \includegraphics[width=\textwidth]{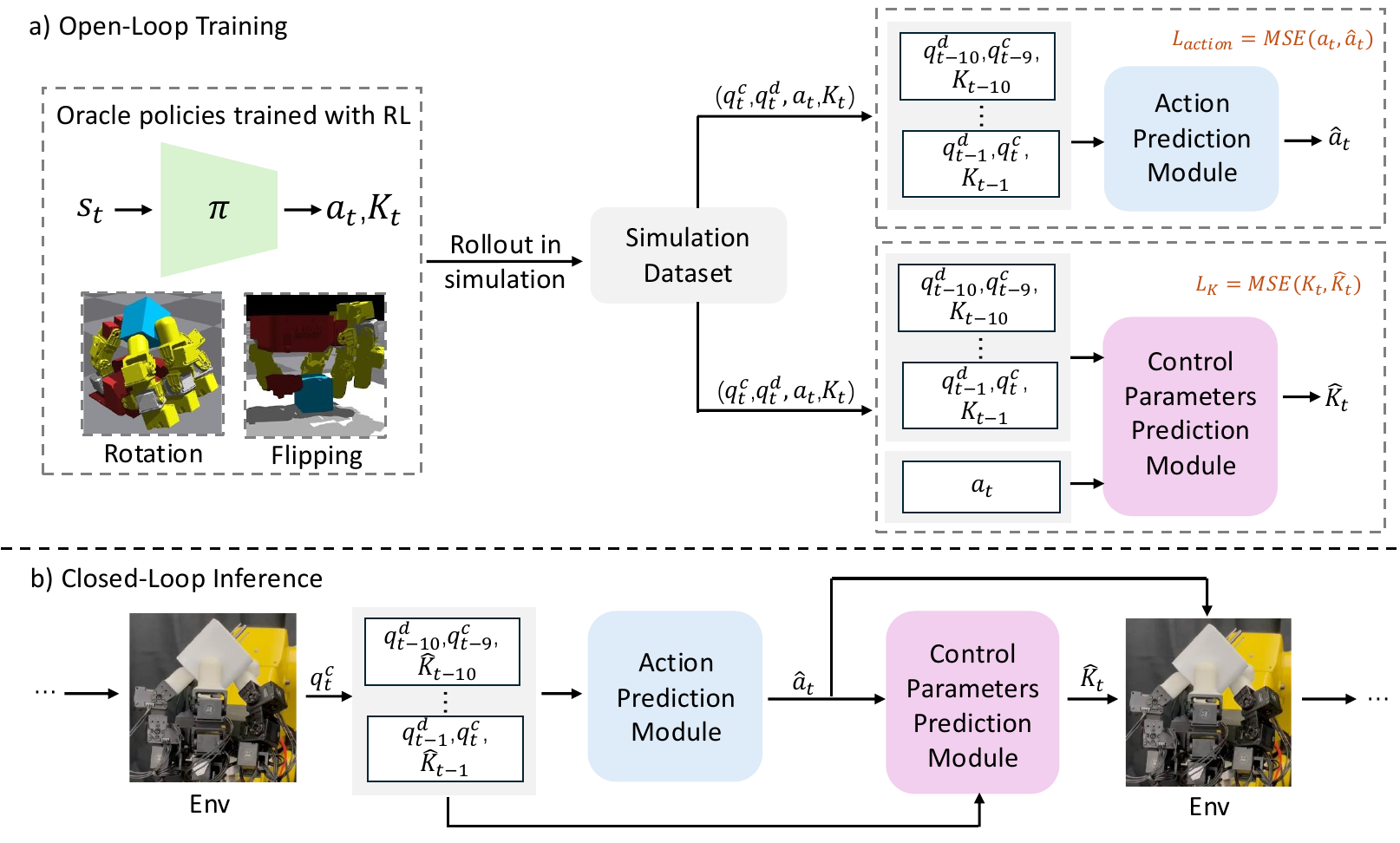}
   \vspace{-20pt}
   \caption{Overview framework of \ours, where $\hat{a}_t$ and $\hat{K}_t$ mean predicted joint position actions and predicted control parameters, respectively.
   % During open-loop training, we first use reinforcement learning to train a base policy and collect sufficient simulation data. Then, we separately train two modules to predict action and control parameters. During closed-loop inference, we leverage the aforementioned two modules and predict motions recursively.
   }
   \label{fig:overview}
   \vspace{-15pt}
\end{figure}

Overall framework of our method is shown in Figure~\ref{fig:overview}.
During training (Figure~\ref{fig:overview} a)), we first collect sufficient data using an oracle policy trained in simulation with diverse object physical parameters.
Then we distill two separate models to predict desired action and control parameters, respectively, based on historical information extracted from the collected dataset.
During inference (Figure~\ref{fig:overview} b)), we recursively predict desired actions for the next step given current and historical observations, and then predict control parameters for the next step based on the generated desired action.
% % version 1
% Here, we decouple the training of trajectories and parameters to reduce learning difficulty, and leverage historical information to substitute for primitive object information in oracle policy. Therefore, by only leveraging simulation data, our method can successfully realize zero-shot sim-to-real transfer and outperforms the previous methods in both simulation and real-world experiments.
% % version 2
By doing so, our method provides two main advantages. First, we include control parameters and actions as part of the observation, which better handles force information and narrows the sim-to-real gap. Second, our method explicitly predicts control parameters with a separate module, which not only adaptively adjusts parameters to improve task performance but also reduces overall training difficulty.

\subsection{Oracle Policy for Data Collection} 
We utilize model-free Proximal Policy Optimization (PPO) reinforcement learning to obtain oracle policies in simulation. Specifically, for each time step $t$, oracle policy $\pi(a_{t}, K_{t} \mid s_{t})$ takes in state $s_{t}$, outputs joint action $a_t$ and $K_t$ simultaneously. The action  $a_t$ is executed using a controller with parameters $K_t$. The desired joint trajectories at $t$ time step are obtained from $q_t^d = q^d_{t-1}+a_t$. The detailed task designs for two contact-rich manipulation tasks are described as follows:

\textbf{State}: States of two tasks $s_t \in \mathbb{R}^{219}$ contain observation of object and robot over the last three time steps. Robot information $s^r_{t} \in \mathbb{R}^{64}$ for each step includes current joint positions $q^c_{t}$, desired joint positions $q^d_{t}$ and controller parameters $K_t$. Object information $s^{obj}_{t} \in \mathbb{R}^{9}$ contains object pose $p^{obj}_{t} \in \mathbb{R}^{6}$ and object property vector $\mu \in \mathbb{R}^{3}$, including scale, mass and friction. In short, the state can be presented as: $s_t \triangleq (s^r_{t-2:t}, s^{obj}_{t-2:t}), 
s^r_{t} \triangleq (q^c_{t}, q^d_{t}, K_t), \ s^{obj}_{t} \triangleq (p^{obj}_{t}, \mu)$

% \begin{gather}
% \label{eq:state}
% s_t \triangleq (s_{r,t-2:t}, s_{o,t-2:t}) \\
% s_{r,t} \triangleq (q_{c,t}, q_{d,t}, K_t), \ s_{o,t} \triangleq (p_{obj,t}, \mu)
% \end{gather}

\textbf{Reward} Reward of two tasks $r_t$ mainly contains four parts: 
% rotation speed reward, contact reward, smoothness reward, and terminate reward: 
\begin{equation}
    r_t = r_{rotation} + r_{contact} + r_{smoothness} + r_{terminate}
\label{eq:reward}
\end{equation}
The rotation speed reward $r_{rotation}$ encourages the object to rotate faster along a certain axis until it reaches the targeted maximum speed. The contact reward $r_{contact}$  encourages binary contacts between the object and the fingertips. The smoothness reward $r_{smoothness}$ penalizes sudden changes of robot joint positions and torques. The terminal reward $r_{terminate}$ penalizes objects when falling off the fingertips (Rotation) and moving too far from the initial positions (Flipping). Details of reward parameters can be found in the appendix. 
% (if there is any?)(TODO: adjust based on flipping policy and detailed reward)

\subsection{Action Prediction and Control Parameters Prediction}
% \subsection{State-based Policy Distillation}
We separately train our student policy into two modules, \textit{i.e.} an action prediction module for trajectory generation and a control parameter prediction module for adapting control parameters, as not only do they encode fundamentally different aspects of the task, but also we want to prevent control parameters prediction from affecting action prediction. 

\textbf{Historical Information for Distillation} Though feasible for task completion, oracle policies cannot be directly transferred to the real world because some primitive information such as object information is not directly accessible in real-world settings. To solve this problem, our method utilizes historical states of robot proprioception to distill primitive information used in oracle policy, enabling the estimation of rotated object properties~\citep{qi2023hand}. Concretely, we use the last ten steps of the current and desired joint trajectories, along with corresponding controller parameters as historical information for policy input.

\textbf{Module Design}
To better leverage historical information, we use self-attention to model temporal historical input for action prediction.
For control parameter prediction, we use cross-attention where current action serves as query and historical input serves as key and value, modeling the relationship between the current action and historical input. 
Although the historical input formulations of the two modules remain the same, their meaning is completely different.
In action prediction, the input mainly indicates the trend of joint trajectory variation, similar to the previous work~\citep{qi2023general, wang2024lessons}.
In control parameter prediction, it mainly indicates an approximate relation between joint actions and control parameters, which can be analogized to how humans infer current control parameters decisions based on historical trajectories and previous parameters.
% \textbf{Network design}

\textbf{Training and Inference} 
Training processes of the two modules are performed in an open-loop manner, meaning all input data is directly retrieved from the collected simulation dataset.
However, during both simulation and real-world inference, our method performs closed-loop behaviors, meaning values of current trajectories are obtained from actual robot sensors.
We linearly map the control parameters from simulation to their corresponding values on the real-world system, with only an approximate estimate of upper and lower bounds instead of careful calculation and tuning.
Furthermore, we find that adding Gaussian noise to the current trajectory values during student policy training is enough for sim-to-real transfer, even though the training dataset is not collected by a teacher policy with input noise randomization. 
This finding indicates that certain randomization on observation can be reduced to ease the teacher policy training procedure.

% \textcolor{blue}{
% \begin{itemize}
%     \item Design.
%     \item why this design
%     \item training details
%     \item illustrate the differences between simulation policy and real-world policy
% \end{itemize}}
%===============================================================================

\section{Experiments}
\label{sec:result}
% \ch{I changed this paragraph, make sure that I didn't change what you want to say}
We conducted experiments in both simulated and real-robot environments to evaluate our proposed method. Specifically, we present results on contact-rich dexterous manipulation tasks, examining two key aspects: in-hand rotation, which emphasizes object-hand interaction, and object flipping, which highlights environment contact. Our investigation primarily addresses key questions through in-hand rotation experiments, with in-hand flipping serving as a supporting task to provide additional insights: \textbf{Q1:} Does our method improve the original oracle performance? \textbf{Q2:} Does our method narrow the sim-to-real gap? \textbf{Q3:} How does our method perform across objects with varying physical parameters? \textbf{Q4:} How do changes in controller parameters impact the results?
% \ch{I think here we don't need itemize, just put in the paragraphs}
% \begin{itemize}
%     \item Does our method improve the original oracle performance?
%     \item Does our method narrows sim-to-real gap?
%     \item How is our method's ability among objects with different physical parameters?
%     % \item What is related to the changes of controller parameters?
%     \item What do controller parameters really change in the manipulation?
%     \item Can our method generalize to different tasks?
% \end{itemize}

\subsection{Experimental Setup}
% \textcolor{blue}{
% \begin{itemize}
%     \item simulation/real world setting
%     \item metrics
%     \item baselines: retrain + only use the trajectory module
% \end{itemize}}
\textbf{Baselines}
We compare our method with two main baselines: 
% \textbf{1) Manual Tuning} Similar to~\citep{chen2023visual, qi2023general}, this baseline trains the oracle policy after carefully tuned control parameters based on trajectories output comparison between simulation and real world. To enhance robustness, a small range of randomization is also added in simulation controller during training. Then, only trajectory prediction module is trained using dataset collected by aforementioned oracle policy. \textbf{2) Ours w/o PD} This baseline trains only the trajectory prediction module using the same dataset collected with our oracle policy. During inference, it predicts only the trajectory action, while keeping the robot controller fixed, similar to the Manual Tuning baseline.
\begin{itemize}[left=0pt]
    \item \textbf{Manual Tuning} Similar to~\citep{chen2023visual, qi2023general}, this baseline trains a new oracle policy with (1) carefully tuned control parameters based on trajectory output comparison between simulation and real world (2) a small range of randomization added on the controller, and then use this new oracle policy for student policy training. Both the new oracle policy and student policy would not output adaptive controller parameters.
    \item \textbf{Ours w/o PD} This baseline replaces the adaptive controller parameter prediction module in the student policy with fixed controller parameters same as manual tuning, and only trains the action prediction module using the dataset collected by our oracle policy.
    % During inference, it predicts only the trajectory action, while keeping the robot controller fixed, similar to the Manual Tuning baseline.
\end{itemize}

\textbf{Metrics}
We quantitatively evaluate the performance of all methods with four metrics~\citep{qi2023hand, qi2023general}:
\begin{itemize}[left=0pt]
    \item \textbf{Rotation Reward/Radians (RotR)} This metric represents the rotation speed of the targeted object around the desired axis. In simulation, it is calculated as the average rotation reward over a trajectory. In the real world, it is calculated by the net rotation of object in radians over a trajectory. 
    \item \textbf{Time To Fail (TTF)} This metric represents the average trajectory length before the object falls off the hand or moves too far from its initial positions in rotation and flipping tasks, respectively.
    \item \textbf{Object linear Velocity (ObjVel)} This metric represents the average magnitude of object linear velocity per action step. It reflects the stability of the targeted object. This metric is only calculated in the simulation because real-world object velocity can not be easily measured in rotation. It is only evaluated in rotation because such behavior is inevitable in flipping.
    \item \textbf{Torque Penalty (Torque)} This metric computes the torque penalty reward per time step to measure the energy cost and it is only measured in simulation.
\end{itemize}

% \subsection{Experimental Results}
% \textcolor{blue}{
% Question:
% \begin{itemize}
%     \item performances differences in simulation?
%     \item sim-to-real improve?
%     \item How's the ability among different masses and frictions of objects?
%     \item whether result's pattern align with assumptions? (Does P actually change related to physical parameters, if not what does it do to make things better, How does D help in this scene, Is it the force that they really discover?)
%     \item control parameters module generalizable? (optional)
% \end{itemize}}

\subsection{Does our Method Improve the Original Oracle Performance?}
\label{sec:q1}
\begin{table*}[b]
\vspace{-20pt}
\begin{minipage}{0.54\textwidth}
\caption{Rotation oracle policy in simulation.
}
\vspace{6pt}
\centering
% \small
\footnotesize 
\setlength{\tabcolsep}{12pt}
\begin{threeparttable}
\resizebox{\columnwidth}{!}{
\begin{tabular}{c|c|c|c|c|c}
\toprule
% \multirow{2}{*}{ Method }   & \multicolumn{2}{c}{ General Set } & \multicolumn{2}{c}{ Hard Set } \\ 
 & Method & RotR $\uparrow$ & TTF $\uparrow$ & Torque $\downarrow$ & ObjVel $\downarrow$ \\
\midrule
% Method & Metrics & Light(44g)  & Medium(94g) & Heavy(160g) & Small & Medium & Large \\ 
\multirow{3}{*}{\makecell{With\\Disturbance}} 
& Manual Tuning & 35.05 & 239.4 & 0.398 & 0.154  \\
& Ours w/o PD & 41.60 & 252.3 & 0.152 & \textbf{0.140}  \\
& Ours & \textbf{43.51} & \textbf{255.9}& \textbf{0.099} & 0.144  \\
   % & Top1 & Top5 & Top1 & Top5  \\  
% \cmidrule(lr){1-1} \cmidrule(lr){2-2} \cmidrule(lr){3-6}
\midrule
% \cmidrule(lr){1-1} \cmidrule(lr){2-3} \cmidrule(lr){4-5}

\multirow{3}{*}{\makecell{Without\\Disturbance}} 
& Manual Tuning & 37.64 & 247.5 & 0.264 & 0.148  \\
& Ours w/o PD & 47.87 & 275.8 & 0.144 & \textbf{0.131}  \\
& Ours & \textbf{52.33} & \textbf{287.7} & \textbf{0.092} & 0.132  \\

\bottomrule
\end{tabular}
}
\end{threeparttable}
\label{table:sim}
% \vspace{-20pt}
\end{minipage}
\hfill
\begin{minipage}{0.45\textwidth}
\caption{Flipping oracle policy in simulation.
}
\vspace{6pt}
\centering
% \small
\footnotesize 
\setlength{\tabcolsep}{12pt}
\begin{threeparttable}
\resizebox{\columnwidth}{!}{
\begin{tabular}{c|c|c|c|c}
\toprule
% \multirow{2}{*}{ Method }   & \multicolumn{2}{c}{ General Set } & \multicolumn{2}{c}{ Hard Set } \\ 
 & Method & RotR $\uparrow$ & TTF $\uparrow$ & Torque $\downarrow$ \\
\midrule
% Method & Metrics & Light(44g)  & Medium(94g) & Heavy(160g) & Small & Medium & Large \\ 
% \multirow{3}{*}{\makecell{With\\Disturbance}} 
% & Manual Tuning & 73.2 & 292.8 & 0.273  \\
% & Ours w/o PD & 73.4 & 294.2 & 0.192  \\
% & Ours & \textbf{133.1} & \textbf{296.2} & \textbf{0.109}  \\
\multirow{3}{*}{\makecell{With\\Disturbance}} 
& Manual Tuning & 91.07 & 295.2 & 0.376  \\
& Ours w/o PD & 82.23 & 295.6 & 0.299  \\
& Ours & \textbf{172.50} & \textbf{296.9} & \textbf{0.140}  \\
   % & Top1 & Top5 & Top1 & Top5  \\  
% \cmidrule(lr){1-1} \cmidrule(lr){2-2} \cmidrule(lr){3-6}
\midrule
% \cmidrule(lr){1-1} \cmidrule(lr){2-3} \cmidrule(lr){4-5}

\multirow{3}{*}{\makecell{Without\\Disturbance}} 
& Manual Tuning & 92.24 & 295.0 & 0.446  \\
& Ours w/o PD & 82.90 & 295.4 & 0.328  \\
& Ours & \textbf{184.00} & \textbf{296.9} & \textbf{0.127}  \\

\bottomrule
\end{tabular}
}
\end{threeparttable}
\label{table:flip}
% \vspace{-20pt}
\end{minipage}
\end{table*}
We first test \ours in a simulation environment with and without randomly applying force disturbances on objects. During validation, we randomize 1024 different initial robot poses and set the simulation controllers exactly the same as those used during training, meaning no controller gap is involved. This experiment aims to show whether performance can be improved by simultaneously adjusting action and control parameters even under the same controllers. Table~\ref{table:sim} and Table~\ref{table:flip} present the quantitative results of baselines and \ours in simulation validation. Compared to the manual tuning baseline, \ours improves performance significantly, especially in RotR and TTF. This demonstrates that our method can effectively stabilize and accelerate the task process with or without disturbance. It's worth noticing that \textit{Ours w/o PD} also reaches relatively good performance, indicating the robustness of trajectories generated by our method. However, in the flipping tasks with results in Table~\ref{table:flip}, \textit{Ours w/o PD} does not outperform the baseline. We believe it is because flipping involves rich contact between both the floor and the dexterous hands, making it more sensitive to controller parameters' variance.
% Furthermore, Figure.~\ref{} demonstrates the training curve of manual tuning and our methods. From visualization, we can see that our method also accelerates the RL training procedure compared to the baseline. 
In conclusion, our method outperforms the baseline in both task performance and training speed, even in the absence of the controller gap.
% \ch{I expect some discussion on why our method outperforms baseline}

\begin{figure}[t!]
% \vspace{-20pt}
\vspace{-10pt}
  \centering
   \includegraphics[width=0.95\textwidth]{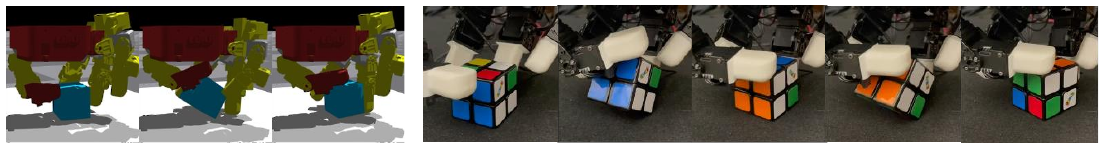}
   \vspace{-10pt}
   \caption{Flipping task performance in simulation and real world.}
   \label{fig:flipping}
   \vspace{-10pt}
\end{figure}

\begin{figure}[t]
  \centering
   \includegraphics[width=0.95\textwidth]{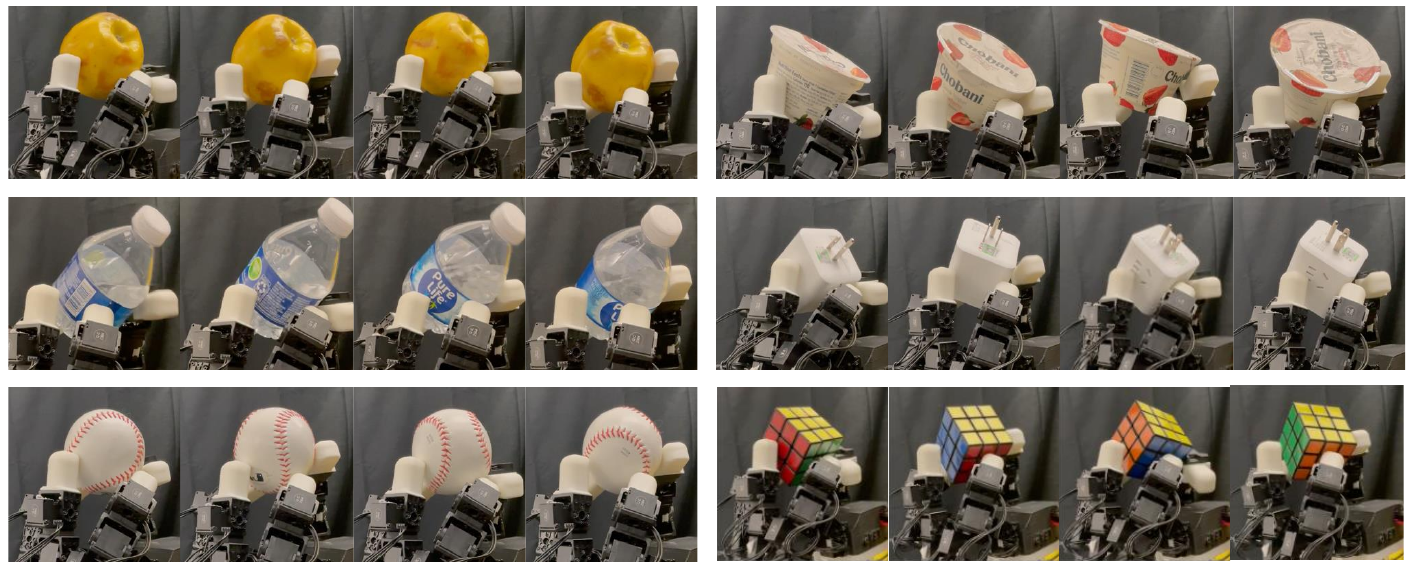}
   \vspace{-10pt}
   \caption{Real-world results of object rotation with different physical parameters. }
   \label{fig:rotation}
   \vspace{-10pt}
\end{figure}

\begin{table*}[t]
\vspace{-10pt}
\caption{Quantitative results of rotation performance for objects with different masses and frictions.
}
\vspace{2pt}
\centering
% \small
\footnotesize 
\setlength{\tabcolsep}{8pt}
\begin{threeparttable}
\resizebox{0.95\textwidth}{!}{
\begin{tabular}{c|c|ccccccc}
\toprule
% \multirow{2}{*}{ Method }   & \multicolumn{2}{c}{ General Set } & \multicolumn{2}{c}{ Hard Set } \\ 
% Method & Metrics & Light(44g)  & Medium(94g) & Heavy(160g) & Small & Medium & Large \\ 
% \cline{3-8}
% \multicolumn{2}{c|}{} & Cube(94g) & Bottle(150g) & Pencil Box(44g) & Apple(221g) & Yogurt(164g) & Baseball(144g) & Overall \\
\multicolumn{2}{c|}{} & Cube(94g) & Bottle(150g) & Apple(221g) & Yogurt(164g) & Baseball(144g) & Average \\
   % & Top1 & Top5 & Top1 & Top5  \\  
% \cmidrule(lr){1-1} \cmidrule(lr){2-2} \cmidrule(lr){3-6}
\midrule
% \cmidrule(lr){1-1} \cmidrule(lr){2-3} \cmidrule(lr){4-5}

\multirow{2}{*}{\makecell{Manual\\Tuning}} 
% & Anyteleop & 36.2\% & 16.8\% & 0.013 & 0.05      \\
& RotR $\uparrow$ & 1.963 & 2.875 & 1.914 & 2.943 & 2.745 &2.431\\
% \cmidrule(lr){1-1} \cmidrule(lr){2-3} \cmidrule(lr){4-5}
& TTF $\uparrow$ & 286.5 & 266.9 & 242.7 & \textbf{300} & 239.6 & 272.4\\
% \cmidrule(lr){1-1} \cmidrule(lr){2-2} \cmidrule(lr){3-6}
\midrule
\multirow{2}{*}{\makecell{Ours\\w/o PD}} 
% & Anyteleop & 34.5\% & 16.8\% & 0.01 &       \\
& RotR $\uparrow$ & 5.498 & 4.285 & 4.492 & 5.424 & 4.681 & 4.986\\
% \cmidrule(lr){1-1} \cmidrule(lr){2-3} \cmidrule(lr){4-5}
& TTF $\uparrow$ & \textbf{297.7} & 281.7 & 291.6 & \textbf{300} & 271.6 & 287.2\\

\midrule
\multirow{2}{*}{Ours} 
% & Anyteleop & 34.5\% & 16.8\% & 0.01 &       \\
& RotR $\uparrow$ & \textbf{9.386} & \textbf{14.006} & \textbf{9.676} & \textbf{15.017} & \textbf{11.342} & \textbf{11.041}\\
% \cmidrule(lr){1-1} \cmidrule(lr){2-3} \cmidrule(lr){4-5}
& TTF $\uparrow$ & 289.3 & \textbf{300} & \textbf{300} & \textbf{300} & \textbf{292.2} & \textbf{292.6}\\

\bottomrule
\end{tabular}
}
\end{threeparttable}
\label{table:rot}
\vspace{-16pt}
\end{table*}

\subsection{Does our Method Narrow the Sim-to-real Gap?}
\label{sec:q2}

We directly deploy \ours in real-world scenarios. For the in-hand rotation task, we use twelve different real-world unseen objects with varying masses and frictions, and validate rotation performance based on the average metrics over ten randomly sampled initial robot poses per object. Table~\ref{table:rot} and Figure~\ref{fig:rotation} present quantitative results and visualizations of \ours along with baselines across different objects, respectively. Compared to the simulation results, the performance gap among different methods is more pronounced in the real world, where \ours significantly outperforms the baseline under zero-shot sim-to-real transfer. Also, \textit{Ours w/o PD} achieves relatively strong performance, demonstrating that trajectories generated by \ours can be more robust when transferred to real-world scenarios. It's worth noticing that the performance gap between \ours and \textit{Ours w/o PD} is much larger in the real world than in simulation. This finding highlights the necessity of adaptively adjusting control parameters at every step in real-world robots, which further emphasizes the importance of adaptive controller prediction in tackling sim-to-real issues. We also conduct real-world experiments on the flipping task with visualizations shown in Figure~\ref{fig:flipping}, demonstrating the real-world task generalizability of our method.

\begin{figure}[t!]
  \centering
   \includegraphics[width=0.9\textwidth]{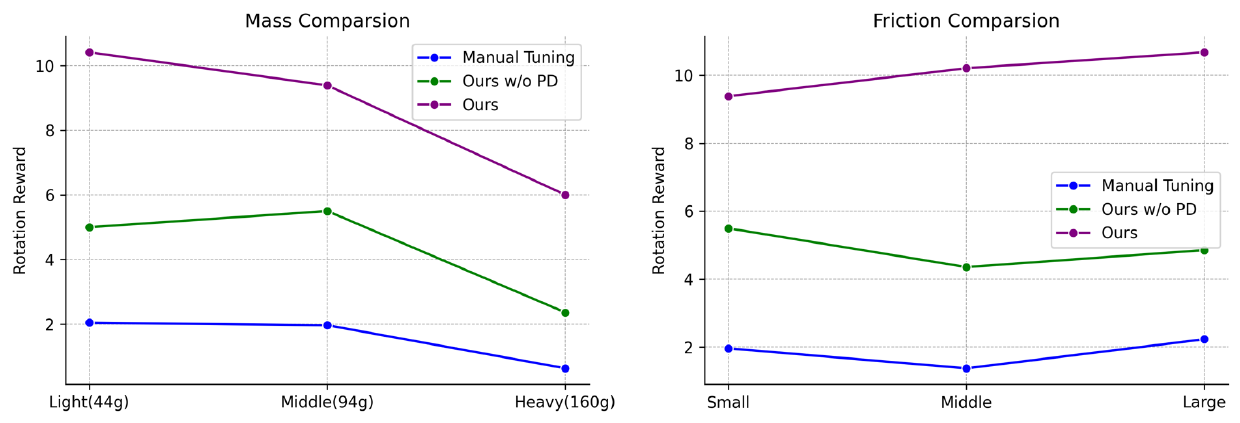}
   \vspace{-10pt}
   \caption{Visualization for same-shape objects rotation with different masses and frictions.}
   \label{fig:mf}
   \vspace{-15pt}
\end{figure}

\begin{table*}[!ht]
% \vspace{6pt}
\caption{Quantitative results for same-shape objects rotation with different masses and frictions.
}
\vspace{2pt}
\centering
% \small
\footnotesize 
\setlength{\tabcolsep}{14pt}
\begin{threeparttable}
\resizebox{0.9\textwidth}{!}{
\begin{tabular}{c|c|ccc|ccc}
\toprule
% \multirow{2}{*}{ Method }   & \multicolumn{2}{c}{ General Set } & \multicolumn{2}{c}{ Hard Set } \\ 
\multirow{2}{*}{Methods} & \multirow{2}{*}{Metrics} & \multicolumn{3}{c|}{Mass} & \multicolumn{3}{c}{Friction}\\
% Method & Metrics & Light(44g)  & Medium(94g) & Heavy(160g) & Small & Medium & Large \\ 
% \cline{3-8}
 &  & Light & Medium & Heavy & Small & Medium & Large \\
   % & Top1 & Top5 & Top1 & Top5  \\  
% \cmidrule(lr){1-1} \cmidrule(lr){2-2} \cmidrule(lr){3-6}
\midrule
% \cmidrule(lr){1-1} \cmidrule(lr){2-3} \cmidrule(lr){4-5}

\multirow{2}{*}{\makecell{Manual\\Tuning}} 
% & Anyteleop & 36.2\% & 16.8\% & 0.013 & 0.05      \\
& RotR $\uparrow$ & 2.042 & 1.963 & 0.628 & 1.963 & 1.374 & 2.231 \\
% \cmidrule(lr){1-1} \cmidrule(lr){2-3} \cmidrule(lr){4-5}
& TTF $\uparrow$ & 286.5 & 286.5 & 243 & 286.5 & 239.9 & 288\\
% \cmidrule(lr){1-1} \cmidrule(lr){2-2} \cmidrule(lr){3-6}
\midrule
\multirow{2}{*}{\makecell{Ours\\w/o PD}} 
% & Anyteleop & 34.5\% & 16.8\% & 0.01 &       \\
& RotR $\uparrow$ & 4.998 & 5.498 & 2.356 & 5.498 & 4.355 & 4.855\\
% \cmidrule(lr){1-1} \cmidrule(lr){2-3} \cmidrule(lr){4-5}
& TTF $\uparrow$ & \textbf{298.8} & \textbf{297.7} & 264.9 & \textbf{297.7} & 278.6 & 286.6\\

\midrule
\multirow{2}{*}{Ours} 
% & Anyteleop & 34.5\% & 16.8\% & 0.01 &       \\
& RotR $\uparrow$ & \textbf{10.414} & \textbf{9.386} & \textbf{5.998} & \textbf{9.386} & \textbf{10.211} & \textbf{10.681}\\
% \cmidrule(lr){1-1} \cmidrule(lr){2-3} \cmidrule(lr){4-5}
& TTF $\uparrow$ & 258.2 & 289.3 & \textbf{300} & 289.3 & \textbf{300} & \textbf{290.4}\\

\bottomrule
\end{tabular}
}
\end{threeparttable}
\label{table:mf}
\vspace{-20pt}
\end{table*}

\subsection{How does our Method Perform across Objects with Varying Physical Parameters?}
\label{sec:q3}

To further evaluate our policy, we perform additional tests on objects with different masses and friction coefficients. To ensure better control over confounding factors, we use a hollow cube for mass experiments and vary its mass by inserting different internal objects, and use cubes of different textures with the same mass for friction experiments. As shown in Table~\ref{table:mf} and Figure~\ref{fig:mf}, our method significantly outperforms both baselines, especially on heavy objects, indicating that it can better adapt to objects with different physical parameters. 
Also, the pattern of our method broadly aligns with our force-based predictions, namely, the lightest and smoothest objects exhibit the highest speed and lowest stability, respectively. 
% It is worth noticing that all methods experience a sudden performance drop in heavy objects, which reveals that mass might be more crucial than the friction.

% \subsection{Can our Method Generalize to Different Tasks?}
% \label{sec:q5}

% % \input{Tables/flipping_table}

% To validate the generalizability of our methods, we also implement \ours on the flipping tasks with the same two baselines. Figure.~\ref{fig:flipping} illustrates task performance in both simulation and real world along with quantitative results in simulation in Table.~\ref{table:flip}. This indicates that our method is suitable among contact-rich dexterous manipulation, which exhibits good potential in wider applications.

% \subsection{What is related to the changes of controller parameters?}
\subsection{ How do Changes in Controller Parameters Impact the Results?}
\label{sec:q4}

\begin{figure}[t]
  \centering
   \includegraphics[width=0.95\textwidth]{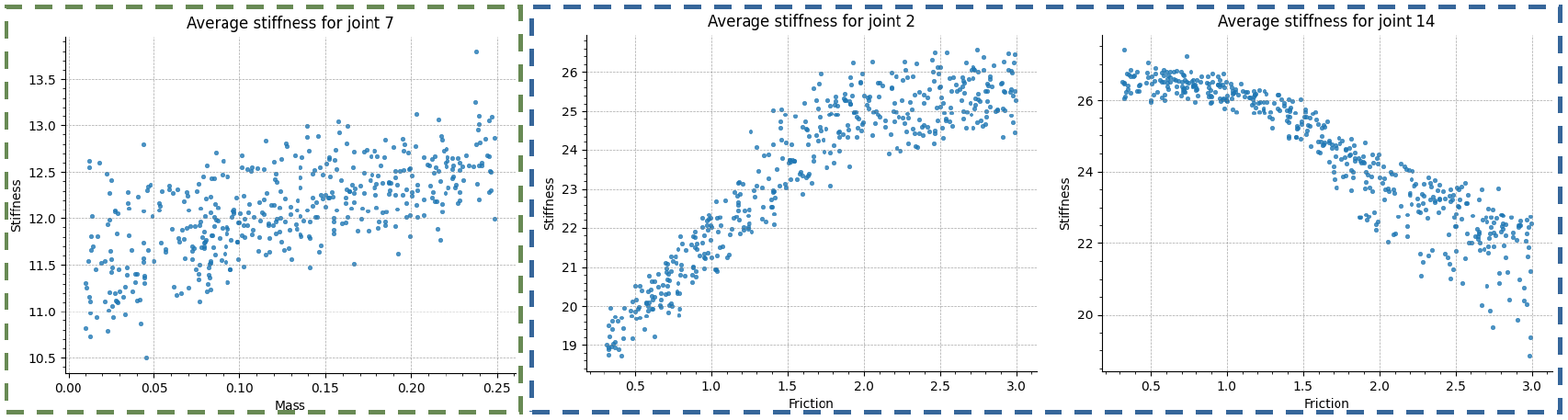}
   \vspace{-8pt}
   \caption{Average stiffness curve under mass (left) and friction change (middle and right).}
   \label{fig:simdataset}
   \vspace{-10pt}
\end{figure}

\begin{figure}[t]
  \centering
   \includegraphics[width=0.95\textwidth]{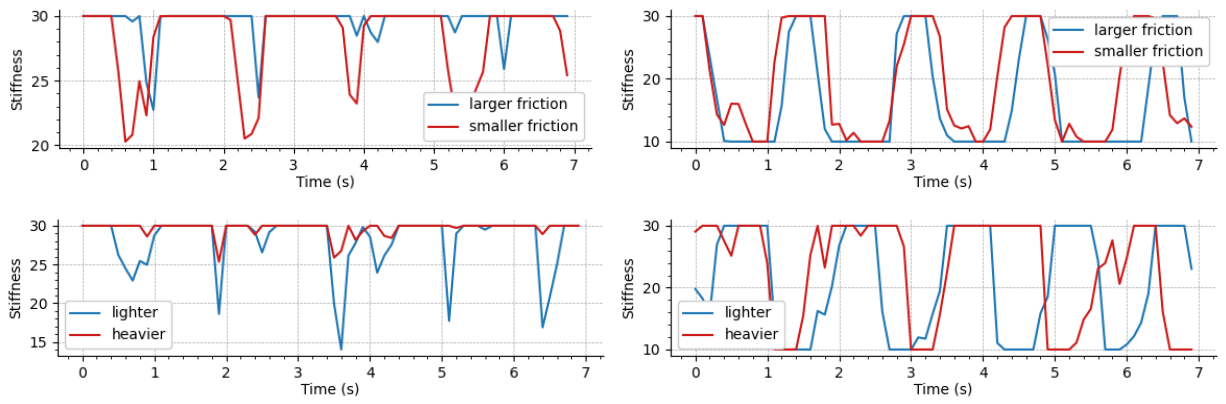}
   \vspace{-10pt}
   \caption{Stiffness over time under varying mass and friction.}
   \label{fig:plot}
   \vspace{-15pt}
\end{figure}

%=================================== Chatgpt =========================================
In this section, we investigate the fundamental question: how do the learned controller parameters affect dexterous manipulation performance? To simplify the analysis, we focus solely on the pattern of learned stiffness, as it contributes more than damping to the task performance. Theoretically, changes in stiffness directly influence the resulting joint torques, thereby modulating the contact forces between the robot and the manipulated object. Given that objects with different physical properties (e.g., mass and friction) require distinct contact force profiles, we hypothesize that variations in stiffness are closely related to object-specific configurations, aiming to provide better adaptation to varying force requirements.

To validate this hypothesis, we first analyze data collected in the simulation. Figure~\ref{fig:simdataset} illustrates the relationship between the average stiffness observed across trajectories and object mass as well as surface friction. The results reveal that stiffness increases monotonically with mass (left of Figure~\ref{fig:simdataset}), consistent with the intuitive physical principle that heavier objects require greater force for manipulation. However, the relationship between stiffness and friction is more nuanced: in some cases, stiffness increases with friction (middle of Figure~\ref{fig:simdataset}), while in others it decreases (right of Figure~\ref{fig:simdataset}). We attribute this inconsistency to task-dependent dynamics. For instance, in rotation tasks, friction may primarily resist angular motion at certain joints, while at others it may act more like a supporting or pushing force. This suggests that the role of friction—and consequently the required stiffness—depends on both the object's properties and the specific joint-task interaction.

To further investigate this relationship in real-world settings, we isolate the controller parameters prediction module and provide ground-truth actions to only predict controller parameters. This allows us to observe the influence of object properties without the confounding effects of policy action noise. Figure~\ref{fig:plot} presents the trends across objects with varying mass and friction. Two consistent patterns emerge: (1) for heavier objects, stiffness tends to increase at specific time steps or remain at its maximum value for longer durations; (2) for smoother objects, stiffness exhibits similar patterns at some joints but opposite trends at others, again indicating task- and joint-specific behavior.

% In summary, our findings demonstrate that the learned stiffness parameter adapts systematically to object mass and friction, which support our hypothesis that variations in controller parameters reflect underlying variations in the required contact forces for effective dexterous manipulation and the learned control parameters benefit performance through better adjustment of different force requirements.
In summary, our results show that the learned stiffness adapts systematically to object mass and friction, validating our hypothesis that controller parameters encode variations in required contact forces and thereby enhance manipulation performance through better adjustment to force requirements.

\section{Related Work}
% \subsection{Sim-to-real Transfer in Dexterous Manipulation}
\textbf{Sim-to-real Transfer in Dexterous Manipulation} Dexterous manipulation tasks typically involve complex interactions between robots and objects through contact~\citep{lin2024twisting,akkaya2019solving,andrychowicz2020learning,teeple2022controlling,yin2023rotating}. Simulation has proven to be an effective way to learn these behaviors~\citep{guo2024phygrasp, zhao2024dexh2r, yang2024anyrotate,lan2023dexcatch}, as teleoperation~\citep{wang2024dexcap, shaw2024bimanual, fu2024mobile,arunachalam2023dexterous,qin2022dexmv,zhao2024aloha} is often not feasible due to the embodiment gap and the delicate nature of the tasks. However, the sim-to-real gap remains a significant challenge~\citep{guo2024phygrasp, xu2023joint}. To bridge this gap, various approaches have been explored, including system identification, policy fine-tuning, and domain randomization~\citep{wang2024lessons,chen2021understanding}. However, previous methods have largely overlooked the sim-to-real controller gap, while our work focuses on narrowing this sim-to-real controller-level gap and significantly improves task performance.

% \subsection{}
\textbf{Learning Adaptive Force Control} Learning adaptive force control has been shown to be beneficial for contact-rich manipulation tasks, as varying control parameters can regulate the robot's behavior during interaction. Several works in the literature demonstrate its effectiveness for various contact-rich tasks, such as robotic table wiping~\citep{martin2019variable, wang2022safe}, object pivoting~\citep{buchli2011learning, zhang2023efficient}, and assembly~\citep{beltran2020variable,zhang2023efficient,zhang2024bridging}. However, this approach has received relatively little attention in the context of dexterous manipulation, and the question of whether and how such a method influences dexterous hand manipulation has not been specifically answered. In our work, we apply the idea of adaptive control to contact-rich dexterous manipulation and prove its efficiency in performance improvement with ample quantitative results, visualization, and discussions.
\label{sec:rw}

% \subsection{Dexterous Manipulation}
% \subsection{Sim-to-real Improvement Force Control}
\section{Conclusion}
\label{sec:conclusion}
% In this work, we address that applying adaptive control parameters in dexterous hands can better narrow the sim-to-real gap, and then propose a novel method that jointly outputs actions and control parameters based on historical information. 
% To validate our method, we conduct comprehensive experiments both in simulation and real-world scenarios and analyze in which way control parameters improve the final performance using in-hand rotation tasks. Also, we implement our method into the flipping task, which proves its generalizability of dexterous manipulation. 
% In the future, we plan to expand the range of our policy so that one control parameter prediction could apply to various dexterous tasks. Also, if hardware supports, another potential direction is to online fine-tune based on force feedback.
In this work, we address the challenge of narrowing the sim-to-real gap in dexterous manipulation by applying adaptive control parameters, and propose a novel method that jointly outputs actions and control parameters based on historical information.
To validate our approach, we conduct comprehensive experiments in both simulation and real-world scenarios, and analyze how control parameters contribute to performance improvements through in-hand rotation tasks. We also apply our method to the flipping task, demonstrating its generalizability across dexterous manipulation settings.
In the future, we plan to expand the applicability of our policy so that multiple dexterous tasks can share a single control parameters prediction module. Additionally, if supported by hardware, another promising direction is to perform online fine-tuning based on real-time force feedback.
%===============================================================================

\clearpage
% The acknowledgments are automatically included only in the final and preprint versions of the paper.
\section{Limitations}
\label{sec:limitation}
Limitations of our method arise primarily in two aspects.
First, our approach currently does not incorporate real-world force or tactile sensing due to hardware limitations. As a result, it relies solely on proprioceptive information, which may not be sufficient to fully recover the system state, particularly in contact-rich scenarios. Future work could explore integrating our method with high-fidelity force feedback to further improve real-world performance.
Second, the real-world evaluation is limited to the LeapHand platform, constrained by the available hardware. In future work, we aim to extend our method to other dexterous hand platforms to assess its generalizability across different robotic embodiments.
% \textcolor{blue}{
% \begin{itemize}
%     \item object shape differences affect the generalization
%     \item lack of online finetune based on the force feedback
% \end{itemize}}

\bibliography{main}  % .bib

\end{document}